# Fooling Vision and Language Models Despite Localization and Attention Mechanism

Xiaojun Xu[1,2], Xinyun Chen[2], Chang Liu[2], Anna Rohrbach[2,3], Trevor Darrell[2], Dawn Song[2]

[1]Shanghai Jiao Tong University, [2]EECS, UC Berkeley, [3]MPI for Informatics

## Abstract

*Adversarial attacks are known to succeed on classifiers, but it has been an open question whether more complex vision systems are vulnerable. In this paper, we study adversarial examples for vision and language models, which incorporate natural language understanding and complex structures such as attention, localization, and modular architectures. In particular, we investigate attacks on a dense captioning model and on two visual question answering (VQA) models. Our evaluation shows that we can generate adversarial examples with a high success rate (i.e., $> 90\%$) for these models. Our work sheds new light on understanding adversarial attacks on vision systems which have a language component and shows that attention, bounding box localization, and compositional internal structures are vulnerable to adversarial attacks. These observations will inform future work towards building effective defenses.*

## 1. Introduction

Machine learning, especially deep learning, has achieved great success in various application scenarios, such as image classification, speech recognition, and machine translation. However, recent studies prove the existence of adversarial examples for many vision-based learning models, which may hinder the adoption of deep learning techniques to security-sensitive applications [19, 43, 56, 65]. Most existing works consider image classification and demonstrate that it is almost always possible to fool these models to classify an adversarially generated image as a class specified by the adversary [66]. Albeit numerous defenses have been proposed [19, 60, 52, 67, 51, 48], almost all of them are later shown to be broken [8, 23, 9].

Recently, there has been an increasing interest in whether adversarial examples are practical enough to attack more complex vision systems [44, 45, 6]. In the latest results of this debate, Lu et al. show that previous adversarial examples constructed to fool CNN-based classifiers cannot fool state-of-the-art detectors [45]. We are interested in whether other forms of localization and/or language context offer effective defense.

In this work, we extend the investigation towards more complex models that not only include a vision component but also a *language component* to deepen our understanding of the practicality of adversarial examples. In particular, we investigate two classes of systems. First, we are interested in dense captioning systems, such as DenseCap [30], which identify regions of interest first and then generate captions for each region. Second, we are interested in visual question answering (VQA) systems, which answer a natural language question based on a given image input. The state-of-the-art VQA systems typically compute attention maps based on the input and then answer the question based on the attended image regions. Therefore, both types of models have a *localization component*, and thus they are good targets for studying whether localization can help prevent adversarial attacks. Further, we explore state-of-the-art VQA models based on Neural Modular Networks [25], and evaluate whether such compositional architectures are also vulnerable to adversarial attacks; in these models, a new network architecture is instantiated for each question type, potentially providing a buffer against attacks.

We evaluate adversarial examples against these vision and language models. We find that in most cases, the attacks can successfully fool the victim models despite their internal localization component via attention heatmaps or region proposals, and/or modular structures. Our study shows that, in an online (non-physical) setting when the attackers have full access to the victim model including its localization component (white-box attack), the generated adversarial examples can fool the entire model regardless of the localization component. Therefore, our evaluation results provide further evidence that employing a localization in combination with a classifier may not be sufficient to defend against adversarial examples, at least in non-physical settings.

We also make the following additional contributions. First, we develop a novel attack approach for VQA models, which significantly outperforms the previous state-of-the-art attacks. Second, we observe and analyze the effect of a *language prior* in attacking VQA models, and define a principle which explains which adversarial examples are likely to fail. In particular, when the target answer is not compatible with the question, it is difficult to find a successful adversarial attack using existing approaches. To sum up, our work sheds new light on understanding adversarial attacks on vision and language systems and shows that attention, bounding box localization and compositional internal structures are vulnerable to adversarial attacks. These observations will inform future work towards building effective defenses.

## 2. Related Work

In the following, we first review recent work on image captioning and visual question answering. We focus on the models that incorporate some form of localization, e.g. soft attention or bounding box detection. We then review the state-of-the-art methods to generate adversarial examples as well as defense strategies against these methods.

**Image Captioning** Most recent image captioning approaches have an encoder-decoder architecture [11, 12, 32, 33, 50, 69]. A spatial attention mechanism for image captioning was first introduced by [73]. They explored soft attention [7] as well as hard attention. Others have adopted this idea [15, 42, 46, 76] or extended it to perform attention over semantic concepts, or attributes [77, 79]. Recently [61] proposed an end-to-end model which regresses a set of image regions and learns to associate caption words to these regions. Notably, [2, 12, 32] exploited object detection responses as input to the captioning system. As opposed to image captioning of the entire image, [30] have proposed *dense captioning*, which requires localization and description of image regions (typically bounding boxes). Some other dense captioning approaches include [40, 74].

**Visual Question Answering.** Early neural models for visual question answering (VQA) were largely inspired by image captioning approaches, e.g. relying on a CNN for image encoding and a RNN for question encoding [17, 49, 62]. Inspired by [73], a large number of works have adopted an attention mechanism for VQA [16, 47, 64, 72, 75, 80]. Semantic attention has been explored by [78]. Other directions explored by recent work include Dynamic Memory Networks (DMN) [36, 71], and dynamic parameter layers (DPP) [55]. Recently a new line of work focused on developing more compositional approaches to VQA, namely neural module networks [3, 4, 25, 29]. These approaches have shown an advantage over prior work for visual question answering which involve complex reasoning.

**Adversarial Examples.** Existing works on adversarial example generation mainly focus on image classification models. Several different approaches have been proposed for generating adversarial examples, including fast gradient-based methods [19, 43], optimization-based methods [66, 10], and others [58, 54]. In particular, Carlini et al. [10] proposed the state-of-the-art attacks under constraints on $L_0$, $L_2$, and $L_\infty$ norms. Our work improves [10] on both attack success rate and adversarial probability.

Another line of research studies adversarial examples against deep neural networks for other tasks, such as recurrent neural networks for text processing [59, 28], deep reinforcement learning models for game playing [41, 26, 34], semantic segmentation [14, 70], and object detection [24, 70]. To our best knowledge, our work is the first to study adversarial examples against vision-language models.

While our work assumes that models are known to the attacker, prior works demonstrate that adversarial examples can transfer between different deep neural networks for image classification [66, 19, 43, 56, 58, 53], which can be used for black-box attacks. We briefly analyze the transferability of VQA models in Appendix D.2.

**Defense against Adversarial Examples.** On the defense side, numerous strategies have been proposed against adversarial examples [19, 60, 52]. Early attempts to build a defense using distillation [60] were soon identified as vulnerable [8]. Some recent proposals attempt to build a *detector* to distinguish adversarial examples from natural images [52, 21, 18, 13]. Others study ensembles of different models and defense strategies to see whether that helps to increase the robustness of deep neural networks [67, 68, 51]. However, He et al. show that with the knowledge of the detector network and the defense strategies being used, an attacker can generate adversarial examples that can mislead the model, while still bypassing the detector [23].

The most promising line of defense strategies is called *adversarial training* [19, 37, 67, 48]. The idea is to generate adaptive adversarial examples and train the model on them iteratively. The latest results along the line [48] show that such an approach can build a robust MNIST model. But the same approach currently fails on extending to CIFAR-10.

## 3. Generating Targeted Adversarial Examples

In this section, we first present a generic adversarial example generation algorithm, and then our implementations for dense captioning models and VQA models.

### 3.1. Background: targeted adversarial examples for a classification model

Consider a classification model $f_\theta(x)$, where $\theta$ is the parameters and $x$ is the input. Given a source image $x$, a targeted adversarial example is defined as $x^\star$ such that

$$f_\theta(x^\star) = y^t \ \wedge \ d(x^\star, x) \leq B \tag{1}$$

where $y^t$ is the target label, and $d(x^\star, x) \leq B$ says that the distance between $x$ and $x^\star$ is bounded by a constant $B$.

Without loss of generality, $f_\theta(x)$ predicts the dimension of the largest softmax output. We denote $J_\theta(x)$ as the softmax output, then a standard training algorithm typically optimizes the empirical loss $\sum_i \mathcal{L}(J_\theta(x_i), y_i)$ with respect to $\theta$ using a gradient decent-based approach. Existing adversarial example generation algorithms leverage the fact that $J_\theta(x)$ is differentiable, and thus solve (1) by optimizing the following objective:

$$\mathbf{argmin}_{x^\star} \mathcal{L}(J_\theta(x^\star), y^t) + \lambda d(x^\star, x) \quad (2)$$

where $\lambda > 0$ is a hyper-parameter. In fact, the state-of-the-art attack [10] approximates the solution to (2) using Adam.

### 3.2. Targeted adversarial examples for DenseCap

The DenseCap model [30] predicts $M = 1000$ regions, ranks them based on confidence, and then generates a caption for each region. It uses a localization network, similar to Fast R-CNN [63], for predicting regions. For each region, the model uses a CNN to compute the embedding and then uses an RNN to generate a sequence of tokens from the embedding to form the caption.

To train the DenseCap model, Johnson et al. include five terms in the loss: four for training the region proposal network, and the last one to train the RNN caption generator. To fool the model to predict the wrong target caption, we can leverage a similar process as discussed above. Note that existing works [24, 70] have demonstrated that an object detection/segmentation model can be fooled by adversarial examples. In this work, we focus on generating adversarial examples to fool the captioning module of the model, while retaining the proposed regions unchanged.

To achieve this goal, assuming the target caption is $C^t$ and the ground truth regions for a source image are $\{R_i\}$, we construct a new set of target region-caption pairs $\{(R_i, C^t)\}$. Using these target region-caption pairs as the new "ground truth", we can use the DenseCap loss, with addition of the $\lambda d(x^\star, x)$ term as in (2), as the new objective, and minimize it with respect to $x^\star$.

### 3.3. Targeted adversarial examples for VQA models

We now briefly present our novel targeted adversarial attack against VQA models. More details can be found in Appendix A. Our design is inspired by two goals: (1) maximizing the probability of the target answer, which is equivalent to the confidence score of the model's prediction; and (2) removing the preference of adversarial examples with smaller distance to the source image, as long as this distance is small enough (i.e., below an upper bound). Our evaluation shows that our algorithm performs better than the previous state-of-the-art [10].

**Algorithm 1** Targeted Adversarial Generation Algorithm against a VQA model

**Input:** $\theta, x, Q, y^t, B, \epsilon, \lambda_1, \lambda_2, \eta, \mathbf{maxitr}$
**Output:** $x^\star$
1     $x^1 \leftarrow x + \delta$ for $\delta$ sampled from a uniform distribution between $[-B, B]$;
2     **for** $i = 1 \rightarrow \mathbf{maxitr}$ **do**
3         $y^p \leftarrow f_\theta(x^i, Q)$;
4         **if** $y^p = y^t$ and $i > 50$ **then**
5             **return** $x^i$ as $x^\star$;
6         $x^{i+1} \leftarrow \mathbf{update}(x^i, \eta, \nabla_x \xi(y^p))$;
7     **return** $x^{\mathbf{maxitr}+1}$ as $x^\star$;

A VQA model takes an additional natural language input $Q$, and predicts an answer from a candidate set of $K$ answers. Similar to (1), a targeted adversarial example $x^\star$ given a question $Q$ is defined to be a solution to:

$$f_\theta(x^\star, Q) = y^t \ \wedge \ d(x^\star, x) \leq B \quad (3)$$

We employ Algorithm 1 to generate the adversarial example $x^\star$. The algorithm takes as input: model parameters $\theta$, source image $x$, question $Q$, target answer $y^t$, the distance bound $B$, and several hyper-parameters: $\epsilon, \lambda_1, \lambda_2, \eta, \mathbf{maxitr}$. This algorithm iteratively approximates the optimal solution to the following objective:

$$\begin{aligned}\xi(y^p) =\ & \mathcal{L}(J_\theta(x^\star, Q), y^t) \\ & + \lambda_1 \cdot \mathbf{1}(y^t \neq y^p) \cdot (\tau - \mathcal{L}(J_\theta(x^\star, Q), y^p)) \\ & + \lambda_2 \cdot \mathrm{ReLU}(d(x^\star, x) - B + \epsilon)\end{aligned} \quad (4)$$

and returns the final result as output. There are two terminating conditions: (1) after at least 50 iterations, if the prediction matches the target, then the algorithm stops and returns the current $x^i$ as output; or (2) after a maximal number of iterations ($\mathbf{maxitr}$), if the prediction still does not match the target, the algorithm returns $x^{\mathbf{maxitr}+1}$ as output.

We now take a closer look at (4). $y^p$ denotes the prediction in each iteration. The objective (4) contains three components. The first is the same as in (2). The second component maximizes the difference between $J_\theta(x, Q)$ and the prediction $y^p$ when $y^p$ is not the target $y^t$. $\tau$ is a constant, e.g., $\log(K)$, set to ensure that the second component is always non-negative. The third component models the constraint $d(x^\star, x) \leq B$ in (3). $\epsilon$ is a small constant set to $(\mathcal{L}(f_\theta(x, Q), y^t) + \lambda_1 \tau)/\lambda_2$ ensures that the adversarial example $x^\star$ which optimizes (4) always satisfies $d(x^\star, x) \leq B$. By using a ReLU function, our attack no longer minimizes the distance $d(x^\star, x)$ if it is smaller than $B - \epsilon$. In practice we choose $d(x, x^\star) = ||x - x^\star||_2/\sqrt{N}$ and set $B = 20$. Other hyper-parameters $\eta, \mathbf{maxitr}$ are the learning rate and the maximal number of iterations. We defer a formal analysis to Appendix A.

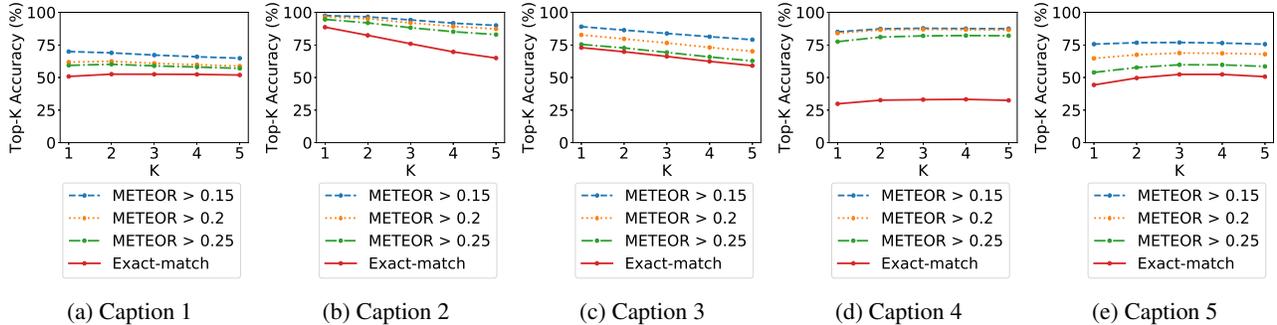

Figure 1: Top-$K$ accuracy on the Caption A dataset averaged across 1000 images generated with each target caption

## 4. Experiments With Dense Captioning

In this section, we evaluate our attacks on Dense-Cap [30], the state-of-the-art dense captioning model. DenseCap employs a region proposal network to first identify the bounding boxes of objects, and then generates captions for each bounding box. We obtain the pre-trained model from their website[1].

To evaluate the attack, we use Visual Genome dataset [35], which was originally used to evaluate Dense-Cap in [30]. For an extensive evaluation, we create the following three attack sets from Visual Genome:

1) Caption A. We randomly select 5 captions as the target captions and 1000 images as the source images;

2) Caption B. We randomly select 1000 captions as target captions and 5 images as source images;

3) Gold. We select 100 images where DenseCap model generates correct captions and manually select target captions irrelevant to the images.

For each caption-image pair, we set the caption as the target, and the image as the source to generate an adversarial example. To evaluate the attack effectiveness, we measure the percentage of top-$K$ predictions from generated adversarial examples that *match* the target captions. We consider two metrics to determine caption matching:

**1) Exact-match.** The two captions are identical.

**2) METEOR$> \omega$.** The METEOR score [38] between the two captions is above a threshold $\omega$. We consider the threshold $\omega$ to be 0.15, 0.2, or 0.25, similar to [30].

Formally, we measure $Acc_{\mu,K}(x^\star, C^t) = \sum_{i=1}^{K} \mu(C^t, C_i)/K$ where $C^t$ is the target caption, $x^\star$ is the adversarial example, $C_i$ for $i = 1, ..., K$ are the top-$K$ predictions for $x^\star$, and $\mu$ is the matching metric (i.e., Exact-match or METEOR$> \omega$).

### 4.1. Results and Observations

The evaluation results on Caption A are presented in Figure 1. Each subfigure shows the results for one target caption. For each caption and each $K \in \{1, 2, 3, 4, 5\}$, we compute $Acc_{\mu,K}$ for each of the 1000 randomly selected images, and report the average value of $Acc_{\mu,K}$ across 1000 images. Each plot contains such 5 top-$K$ accuracy values for each metric described above (see the legend).

[1] https://github.com/jcjohnson/densecap

We observe that using the metric derived from METEOR score, the accuracy is higher than using the Exact-match metric. This is intuitive, since Exact-match is an over-conservative metric, which may treat a semantically correct caption as a wrong answer. In contrast, using METEOR score as the metric can mitigate this issue. Even with Exact-match, we observe that all captions have an average top-$K$ accuracy above $30\%$. Further, for target captions Caption 1-3, the top-1 accuracy is always above $50\%$. That means, at least 500 generated adversarial examples can successfully fool the DenseCap system to produce the exact target captions with the highest confidence score.

We further investigate the number of attack "failures" among caption-image pairs in Caption A. The attack *fails* if none of the top-5 predictions matches the target based on METEOR$> 0.15$. We find only 17 such caption-image pairs, i.e., $0.35\%$ of the entire set, which lead to adversarial attack failure. This means that for the rest $99.65\%$ caption-image pairs, the attacks are successful in the sense that there exists at least one prediction for each adversarial example that matches the target caption. The 17 cases can be found in Appendix B.

The results on Caption B set are similar, and we observe that $97.24\%$ of the caption-image pairs can be successfully fooled in the sense described above. For the Gold set we find that our attack fails only on one image. Due to space limitations, we defer detailed results on Caption B and Gold sets to Appendix B.

Note that the attack does not achieve a 100% success rate. We attribute it to two reasons: (1) it is challenging to train an RNN-based caption generation model to generate the exactly matching captions; and (2) the DenseCap network involves randomness, and thus may not produce the same results for all runs. Still, we observe that the attack success rate is over $97\%$, and thus we conclude that the DenseCap model can be fooled by adversarial examples.

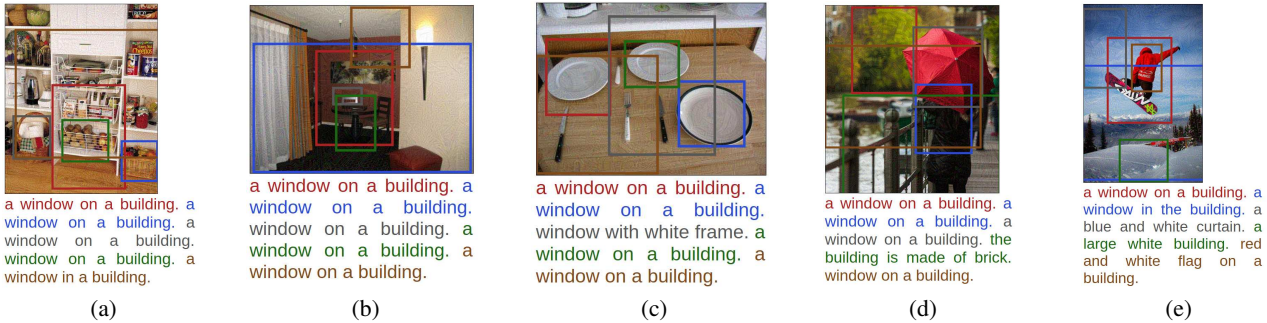

Figure 2: Adversarial examples generated from different images with the target caption to be "a window on a building".

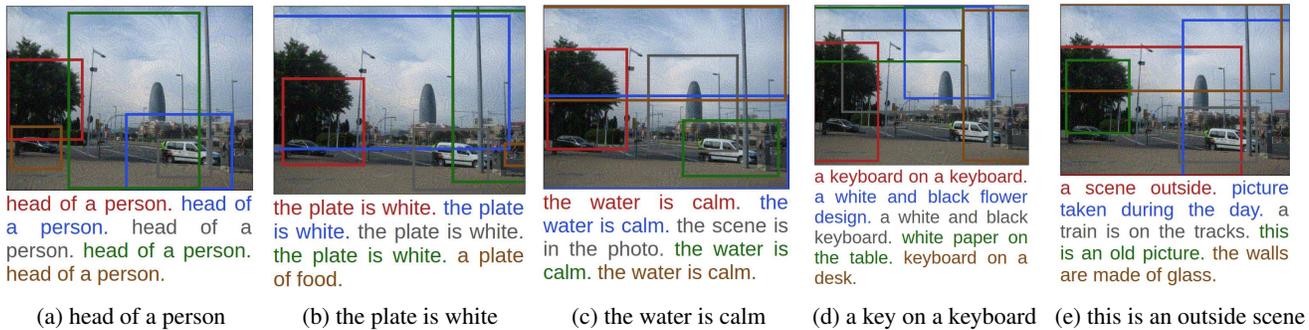

Figure 3: Adversarial examples generated from Image 4 with different target captions (shown as sub-figure captions).

### 4.2. Qualitative Study

We conduct qualitative study to investigate the generated adversarial examples and their predictions. In Figure 2, we present five adversarial examples generated for the same target caption. We see that most of the predicted captions exactly match the target (e.g., all top-5 predictions for Figure 2a and Figure 2b), or be semantically equivalent to the target (e.g., the top-2 prediction for Figure 2e). We further examine the bounding boxes of the regions proposed by the model. We find that the model localizes objects in the adversarial examples, although the caption generation module of the model is completely fooled. For example, in Figure 2c, the model can successfully identify the plates, but label all of them as "a window on a building".

To further understand this effect, in Figure 3, we show the adversarial examples generated from the same source image but with different target captions. We observe that all adversarial images look identical to each other, and the regions proposed for different images are also similar. For example, we observe that the top proposed regions for the first four images all circumscribe the tree on the left. However, the top captions generated for this region are all different, and match the target captions very well.

## 5. Experiments with VQA

In this section, we evaluate the previous state-of-the-art attack [10] and our novel algorithm on two VQA models.

We also investigate the effect of adversarial attacks on *attention maps* of the VQA models to gain more insights about the way the attacks work. Finally, we analyze the successes and failures of our attacks with respect to *language prior*. More results on qualitative study, transferability, and further investigations to the failure cases can be found in Appendix D and E.

### 5.1. Models

We experiment with two state-of-the-art models for open-ended visual question answering, namely the MCB model [16], which is the winner of the VQA challenge in 2016, and the compositional model N2NMN [25]. Both models achieve similar performance on the VQA benchmark [5], while being very different in terms of internal structures. MCB relies on a single monolithic network architecture for all questions, while N2NMN dynamically predicts a network layout for every given question. In our experiments we investigate whether such compositional dynamic architecture is more resilient than the monolithic one.

We retrieve the pre-trained model of MCB from their website[2], and the pre-trained model of N2NMN by contacting the authors through email directly. The code implementing N2NMN is acquired from the website.[3] Notice that the MCB model is trained not only on the VQA dataset but also on the Visual Genome dataset [35], while the N2NMN model only considers the VQA dataset.

---
[2] https://github.com/akirafukui/vqa-mcb
[3] https://github.com/ronghanghu/n2nmn

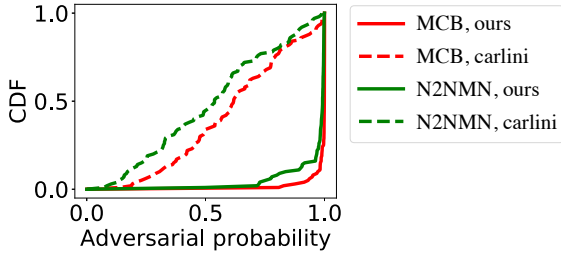

Figure 4: CDF of adversarial probability on the Gold set.

| Image # | | 1 | 2 | 3 | 4 | 5 |
|---|---|---|---|---|---|---|
| MCB | ours | 94.67 | 94.78 | 94.97 | 95.02 | 95.15 |
| | CW [10] | 94.10 | 94.28 | 94.27 | 94.52 | 94.78 |
| N2NMN | ours | 94.25 | 94.53 | 95.57 | 95.80 | 96.15 |
| | CW [10] | 93.82 | 93.78 | 95.02 | 95.08 | 95.37 |

Table 1: Attack success rate (%) on VQA-A.

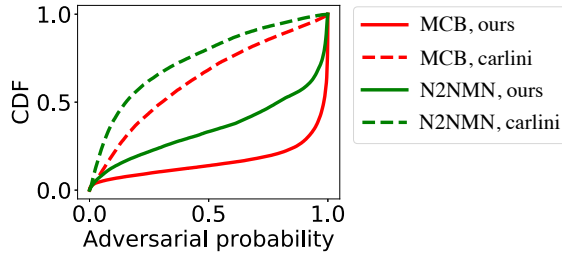

Figure 5: CDF of adversarial probability on VQA-A.

### 5.2. Datasets

To evaluate different adversarial example generation algorithms we derive three datasets from the VQA dataset [5]. In particular, we choose source images and question-answer targets from the VQA validation set as follows:

1) VQA-A: We randomly select 6,000 question-answer pairs and 5 source images to constitute 30,000 triples;

2) VQA-B: We randomly select 5,000 source images and 10 question-answer pairs to construct 50,000 triples;

3) Gold: We manually select 100 triples, such that MCB and N2NMN models can correctly answer the questions based on the images, and the target answers are plausible for the questions but incorrect for the images.

For each triple of question-answer-image, we generate an adversarial example close to the source image using the answer as the target. More details can be found in Appendix C.

### 5.3. Evaluation metrics

Given a set of question-answer pairs $(Q, y^t)$ and the generated adversarial examples $\{x^\star\}$, we evaluate two metrics: the *attack success rate* and *the adversarial probability*.

**Attack success rate.** The attack is considered successful if $f_\theta(x^\star, Q) = y^t$. The attack success rate is computed as the percentage of successful attacks over all triples in a dataset.

**Adversarial probability.** The *adversarial probability* is computed as $J_\theta(x^\star, Q)_{y^t}$, where $J(\cdot, \cdot)_i$ indicates the $i$-th dimension of the softmax output. Adversarial probability indicates the confidence score of the model to predict the target answer $y^t$, and thus provides a fine-grained metric.

### 5.4. Results

Here we report the overall success of adversarial attacks on VQA models, and also compare our new algorithm described above with the performance of the previous attack algorithm (CW [10]) applied to this novel VQA setting. We present the quantitative results below, and defer more qualitative results to Appendix D.1.

**Gold.** For both MCB and N2NMN models, we achieve 100% attack success rate using either approach. Note that both models can correctly answer all the questions on the original source images. The 100% attack success rate for both VQA models shows that both of them are vulnerable to targeted adversarial examples.

We inspect the adversarial probabilities of the generated adversarial examples, and plot the Cumulative Distribution Function (CDF) in Figure 4. Note that a lower CDF curve indicates a higher probability in general. From the figure we observe that the CDF curve of N2NMN is above MCB's, indicating that N2NMN is slightly more resilient than MCB. However, we also observe that for both models, almost in all cases the adversarial probability is above 0.7. Thus, we conclude that our attack is very successful at misleading the VQA models to predict the target answers. We also observe that the CDF curve of CW attack is much higher than ours, showing that our approach is more effective at achieving a high adversarial probability. Overall, we show that such attacks can be performed very successfully for target answers that are meaningful to questions.

**VQA-A.** We further investigate VQA adversarial examples across a wide range of target question-answer pairs. We separately compute the attack success rate using each image as the source. The results are presented in Table 1, and the corresponding CDF curves are plotted in Figure 5. We can draw similar conclusions as for the Gold set: (1) the attack success rate is high, i.e., $> 90\%$; (2) the adversarial probability of our attack is high; and (3) our attack is more effective than CW attack.

We observe that the attack's performance against N2NMN model is worse than against MCB. In particular, from Figure 5, we see that the adversarial probability of at-tacks generated on the N2NMN model is significantly lower than the MCB model. This further shows that N2NMN model is somewhat more robust against adversarial attacks. We also observe that the attack success rate with respect to different images does not vary too much. We hypothesize that the attack success is not sensitive to a source image, but more dependent on a target question-answer pair. Our further investigations on VQA-B and language priors below provide more evidence to confirm this hypothesis.

| Original image | Benign attention maps | | Adversarial attention maps | |
| --- | --- | --- | --- | --- |
| | MCB Attention | N2NMN Attention | MCB Attention | N2NMN Attention |
| **What is the man holding?** 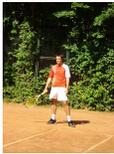 | Original answer: **racket** | | Target: **phone** | |
| **What does this sign say?** 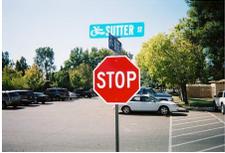 | Original answer: **stop** | | Target: **one way** | |
| **What type of vehicle is this?** 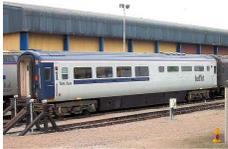 | Original answer: **train** | | Target: **bus** | |

Table 2: Attention maps of benign and adversarial images on MCB and N2NMN models.

The attack success rate is not 100%, which shows that there exist a few question-answer pairs where neither ours nor the CW attack can succeed. In fact, for these question-answer pairs, we have also tried other attack methods and none of them can succeed in fooling the victim VQA model. We find that these question-answer pairs tend to appear infrequently in the training set, and this observation leads to our hypothesis regarding *language prior*. We present more analysis of the language prior in the following section.

**VQA-B.** We test the hypothesis that *the attack success rate is not strongly dependent on the choice of source images* using the VQA-B dataset. In our evaluation, we observe that for 9 out of 10 question-answer pairs, the adversarial examples generated from any of the 5,000 source images fool the victim model with 100% attack success rate. For the one remaining question-answer pair, however, we cannot generate successful adversarial examples from *any* of the source images. This result further confirms our hypothesis. Interestingly, we observe that the "hard" question-answer pairs for the two VQA models are different. For the MCB model, the question is "Why is the girl standing in the middle of the room with an object in each hand?" with the target answer "playing wii"; for the N2NMN model, the question and answer are "Who manufactured this plane?" and "japan", respectively. This suggests that the hard question-answer pairs are model-specific, which further motivates us to investigate language prior in VQA models.

### 5.5. Adversarial examples fool attention mechanism

We conduct a qualitative study to gain more insights as to how the attack succeeds. In the following we use the Gold dataset. In particular, both models in our experiments have attention mechanism. That is, to answer a question, a model first computes an *attention map*, which is a weight distribution over local features extracted from a CNN based on the image and the question. Intuitively, a well-performing model should put more weight, i.e. attend to, the image region that is most informative to answer the question.

We demonstrate the attention heatmaps for three source images and their adversarial counterparts in Table 2. We observe that the adversarial examples mislead the VQA models to ignore the regions that support the correct answer to the question. For example, in the second source image both

MCB and N2NMN focus on the stop sign when answering the question. The adversarial examples fool MCB and N2NMN to pay attention to the street sign instead, which leads to predicting a one-way traffic sign, likely because both signs are long rectangular metal plates. In the last example, the attention is mislead to ignore the rail tracks but focusing on the windows which look similar to those on a bus. Therefore, we observe that adversarial examples can fool both the attention and the classification component of the VQA models to achieve the malicious goal.

### 5.6. Language Prior

We illustrate the *language prior* phenomenon in Figure 6. It provides an example which cannot be successfully attacked by any algorithm in our evaluation. We show an adversarial example generated by our attack algorithm, and the top-5 predictions from the MCB model. Clearly, the model is *confused* about the image and the question. The answer with the highest probability only has a probability of less than $5\%$. Although the model is confused, it cannot be *fooled* to predict the target answer "partly" to the question "what animal is next to the man?". This observation is different than those reported in the literature [66], i.e., that targeted adversarial examples can always be successfully generated against an *image classifier* regardless of the image and the target label. We believe that the observed phenomenon is due to the internal mechanism of a VQA model which learns to process natural language questions and predict semantically relevant answers.

In all previous experiments we choose question-answer pairs from the VQA validation set, and thus the answers are likely meaningful to the questions. To evaluate the effect of language prior we construct the *Non-Sense* dataset. Specifically, we choose question-answer pairs, such that answers do not match the questions semantically, as they belong to questions of a different type (e.g. "what color" vs. "how many"). We find that the attack success rates using our approach against MCB and N2NMN are only $7.8\%$ and $4.6\%$ respectively; the corresponding numbers for CW attack are even lower, $6.8\%$ and $3.8\%$. This experiment further confirms the significance of the language prior.

Prior work has noted the effect of language prior, i.e. that the VQA models capture the training data biases and tend to predict the most frequent answers [1, 20, 27, 31]. We find that N2NMN is more influenced by language prior than MCB. Specifically, N2NMN produces a smaller number of distinct answers, predicting question-relevant answers independent of image content. This may explain why it is more difficult to achieve a high probability on some targets with N2NMN than with MCB. We include more results and analysis in Appendix E.

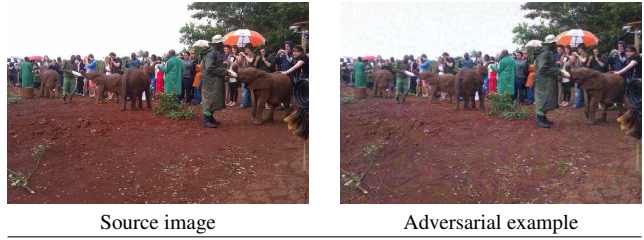

| Rank | Answer | Probability |
|------|--------|-------------|
| 1 | yes | 0.042 |
| 2 | middle | 0.041 |
| 3 | on wall | 0.040 |
| 4 | left | 0.031 |
| 5 | background | 0.025 |

Figure 6: The effect of language prior. The target question / answer are "*What animal is next to the man?*"/"*partly*". We show the top-5 predictions from MCB after the attack.

## 6. Conclusion

In this work, we study adversarial attacks against vision and language models, specifically, dense captioning and visual question answering models. The models in our study are more complex than previously studied image classification models, in the sense that they contain language generation component, localization, attention mechanism, and/or compositional internal structures. Our investigation shows that (1) we can generate targeted adversarial examples against all victim models in our study with a high success rate (i.e., $> 90\%$); and (2) the attacks can either retain the localization output or also fool the attention heatmaps to fool the victim model. While studying attacks on VQA models, as additional contributions, we propose a better attack method than the previous state-of-the-art approach. Also, we observe and evaluate the effect of language prior that may explain which question-answer pairs represent harder targets. Our work sheds new light on understanding adversarial attacks on complex vision and language systems, and these observations will inform future directions towards building effective defenses.

### Acknowledgement

This work was supported in part by the National Science Foundation under Grant No. TWC-1409915, Berkeley DeepDrive, DARPA STAC under Grant No. FA8750-15-2-0104, and Center for Long-Term Cybersecurity. Any opinions, findings, and conclusions or recommendations expressed in this material are those of the author(s) and do not necessarily reflect the views of the National Science Foundation.

# A. Implementation details of improved attacks to VQA models

In this appendix, we detail our attack strategy against VQA models, which is briefly sketched in Section 3.3. We first provide in A.1 the full background setup and some terminology, as well as a formal definition of targeted adversarial examples for VQA models. Then, we present the attack details in A.2 and A.3 using the terminology defined in A.1. In A.4, we will explain other implementation details, such as hyperparameter values, used in our evaluation. In the end, we provide some proofs to the theoretical analysis behind our technique in A.5.

## A.1. Targeted adversarial examples for VQA

We denote a VQA model as $f_\theta(I, Q)$, where $\theta$ is the parameters of the model, $I$ is the input image, and $Q$ is the input question. The output $f_\theta(I, Q)$ is the predicted answer to the question $Q$ given the image $I$.

Most existing VQA models consider this task as a classification problem. That is, they choose the most probable answer among the *top-K most frequent answers* in the training set (or both training and testing set). Typically, state-of-the-art VQA models use $K = 3000$.

We consider that the target to a VQA model $f_\theta$ is a question-answer pair $(Q^{\mathbf{target}}, A^{\mathbf{target}})$, and a targeted adversarial example is an image $I^{\mathbf{adv}}$ such that

$$f_\theta(I^{\mathbf{adv}}, Q^{\mathbf{target}}) = A^{\mathbf{target}} \text{ s.t. } d(I^{\mathbf{adv}}, I^{\mathbf{ori}}) \leq B$$

where we refer to $I^{\mathbf{ori}}$ as the *benign image*.

A neural network-based VQA model $f_\theta$ can also be represented as $f_\theta(I, Q) = \mathbf{argmax}_i J_\theta(I, Q)$, where $J_\theta(I, Q)$ outputs a $K$-dimensional vector in which each dimension indicates the probability of corresponding choice to be the predicted answer. Therefore, we can generate adversarial examples by solving the following optimization problem

$$\mathbf{argmin}_{I^{\mathbf{adv}}} \mathcal{L}(J_\theta(I^{\mathbf{adv}}, Q^{\mathbf{target}}), A^{\mathbf{target}}) \quad (5)$$
$$\text{s.t. } d(I^{\mathbf{adv}}, I^{\mathbf{ori}}) \leq B \quad (6)$$

where $I^{\mathbf{ori}}$ is a benign image, and the goal is to find an adversarial example $I^{\mathbf{adv}}$ that is close to $I^{\mathbf{ori}}$. Typically, $\mathcal{L}$ is chosen as the same loss function for training the model, but other alternatives which are monotonic to the training loss can also be used. In particular, Carlini *et al*. show that the choice of different loss functions has a significant impact on the attack success rate [10], when the attacks are evaluated on MNIST dataset [39]. In this work, we consider $\mathcal{L}$ to be the cross-entropy loss, which is equivalent to the best loss function used in [10].

## A.2. Solving the optimization problem

In our attack method, we approximate the optimization problem using an alternative objective function (4). We restate it below using the notation defined in A.1:

$$\begin{aligned}
\xi(A^{\mathbf{predict}}) &= \mathcal{L}(J_\theta(x, Q^{\mathbf{target}}), A^{\mathbf{target}}) \\
&+ \lambda_1 \cdot \mathbf{1}(A^{\mathbf{target}} \neq A^{\mathbf{predict}}) \\
&\quad \cdot (\tau - \mathcal{L}(J_\theta(x, Q^{\mathbf{target}}), A^{\mathbf{predict}})) \\
&+ \lambda_2 \cdot \mathrm{ReLU}(d(x, I^{\mathbf{ori}}) - B + \epsilon) \quad (7)
\end{aligned}$$

In this formula, we use $x$ to represent the image. Thus the adversarial example is the value of $x$ that minimizes the objective (7). This objective has three components. The first component, $\mathcal{L}(J_\theta(x, Q^{\mathbf{target}}), A^{\mathbf{target}})$, is the same as objective (5). The latter two are the innovations in this work, and we elaborate their design in the following.

**The second component.** The second component is

$$\lambda_1 \cdot \mathbf{1}(A^{\mathbf{target}} \neq A^{\mathbf{predict}}) \cdot (\tau - \mathcal{L}(J_\theta(x, Q^{\mathbf{target}}), A^{\mathbf{predict}}))$$

Here the hyperparameter $\lambda_1$ is used to balance this component and others, and $A^{\mathbf{predict}}$ is the prediction of the original image. The value of $A^{\mathbf{predict}}$ is set dynamically during the iterative optimization process, so that each iteration may choose a different value of $A^{\mathbf{predict}}$. We will explain this process in more details in the next subsection. We set $\tau$ to be a constant, e.g., $\log(K)$ when $\mathcal{L}$ is chosen as the cross-entropy loss. This constant guarantees the second term is always non-negative, especially when $\mathbf{1}(A^{\mathbf{target}} \neq A^{\mathbf{predict}})$. In fact, we have the following theorem:

**Theorem 1.** *Assuming $\tau = \log K$, where $K$ is the number of output classes, $\mathcal{L}$ is the cross-entropy loss, i.e., $\mathcal{L}(u, i) = -\log u_i$, the last layer of $J$ is a softmax operator, and $A^{\mathbf{predict}}$ is the prediction of the model over input image $x$ and question $Q^{\mathbf{target}}$, i.e., $\mathbf{argmax}_i J_\theta(x, Q^{\mathbf{target}})$, then we have*

$$\begin{aligned}
\mathbf{1}(A^{\mathbf{target}} &\neq A^{\mathbf{predict}}) \\
\cdot (\tau - \mathcal{L}(J_\theta(x, Q^{\mathbf{target}}), A^{\mathbf{predict}})) &\geq 0 \quad (8)
\end{aligned}$$

To understand how this component works, we consider two possible cases. First, in the case $A^{\mathbf{predict}} = A^{\mathbf{target}}$, the image generated in the last iteration is already an adversarial example, and thus this component is 0 since $\mathbf{1}(A^{\mathbf{target}} \neq A^{\mathbf{predict}}) = 0$. In this case, optimizing objective (7) is equivalent to maximizing the probability of predicting the target answer $A^{\mathbf{target}}$.

Second, when $A^{\mathbf{predict}} \neq A^{\mathbf{target}}$, minimizing the second component is essentially maximizing $\mathcal{L}(J_\theta(x, Q^{\mathbf{target}}), A^{\mathbf{predict}})$, which is equivalent to minimizing the probability of the model to predict $A^{\mathbf{predict}}$, which is different from the target answer $A^{\mathbf{target}}$. As for

**Algorithm 2** Targeted Adversarial Generation Algorithm

**Input:** $\theta, I^{\text{ori}}, Q^{\text{target}}, A^{\text{target}}, B, \epsilon, \lambda_1, \lambda_2, \eta, \textbf{maxitr}$
**Output:** $I^{\text{adv}}$

1    $I^1 \leftarrow I^{\text{ori}} + \delta$ for $\delta$ sampled from a uniform distribution between $[-B, B]$;
2    **for** $i = 1 \to \textbf{maxitr}$ **do**
3        $A^{\text{predict}} = f_\theta(I^i, Q^{\text{target}})$;
4        **if** $A^{\text{predict}} = A^{\text{target}}$ and $i > 50$ **then**
5            **return** $I^i$
6        $I^{i+1} \leftarrow \textbf{update}(I^i, \eta, \nabla_x \xi(A^{\text{predict}}))$;
7    **return** $I^{\textbf{maxitr}+1}$

---

the value of the hyperparameter $\lambda_1$, which is used to balance between this component and others, we find that setting $\lambda_1 = 1$ works well in most cases. Notice that in this case, jointly optimizing the first and the second component is equivalent to optimizing the best loss function used in Carlini's attack.

**The third component.** The third component is set to enforce the constraint (6). In particular, $\text{ReLU}(x) = \max(0, x)$ is the rectifier function, and $\epsilon$ is a small positive hyper-parameter that we will explain later. When $d(I^{\text{adv}}, I^{\text{ori}}) \leq B - \epsilon < B$, i.e., constraint (6) is satisfied, the third component is 0, and thus has no effective on the objective. On the other hand, if an adversarial example $I^{\text{adv}}$ does not satisfy constraint (6), we show that it is never optimal for (7) when $\lambda_2 \epsilon$ is large enough. We have the following theorem:

**Theorem 2.** *When $\lambda_2 \epsilon > \mathcal{L}(f_\theta(I^{\text{ori}}, Q^{\text{target}}), A^{\text{target}}) + \lambda_1 \tau$, the solution $I^{\text{adv}}$ minimizing the objective (7) satisfies constraint (6) as well.*

In practice, we can set $\epsilon$ to be a small value (e.g., 2), and set $\lambda_2$ to be a large value (e.g., 10), then the generated adversarial examples end up not activating the ReLU function (i.e., the output of the function is 0). Even when the ReLU function is activated, its value is not larger than $\epsilon$, and thus the constraint (6) is still satisfied.

Notice that in most previous iterative optimization-based approaches [10, 43], optimizing (5) while satisfying constraint (6) is converted into a joint optimization problem of $\mathcal{L}(...) + \lambda d(...)$, which minimizes both the lost function (5) and the distance function $d(I^{\text{adv}}, I^{\text{ori}})$. The most prominent difference is that our approach does not minimize this distance as long as it is within the bound $B$.

### A.3. Putting everything together

The overall adversarial generation method is presented in Algorithm 1 (see the main paper). We restate the algorithm in Algorithm 2 using the notation defined in A.1, and explain the details below.

This algorithm takes the hyper-parameters defined above, along with $\eta$, representing the learning rate, and **maxitr**, representing the maximal number iterations that the algorithm runs. In the algorithm, $I^1$ is initialized with a random starting point satisfying constraint (6) (line 1). Then the algorithm iteratively updates $I^i$ (lines 2-6). In each iteration, the prediction $A^{\text{predict}}$ is first computed (line 3). If this prediction already matches the target, and the algorithm has run for at least 50 iterations, the algorithm stops and returns $I^i$ as a successful adversarial example (lines 4-5). Here, 50 is a hyperparameter that can be further tuned. In this work, we fix it to be 50 in all experiments. On the other hand, if the algorithm does not stop at line 5, then $I^{i+1}$ will be updated based on the gradient $\nabla_x \xi(A^{\text{predict}})$ and the learning rate $\eta$ (line 6). Here, **update** can be any optimization algorithm. We evaluated the algorithm's performance by using SGD, Adam, or RMSProp, and found that Adam always yields the best attack success rate. Therefore, we use Adam as the **update** function through out this work. In the end, if it does not return at line 5 during some iteration, then the algorithm fails at finding an adversarial example, and it returns $I^{\textbf{maxitr}+1}$ as a result. In our evaluation, we set $\eta = 1.0$ and $\textbf{maxitr} = 1000$ for evaluation.

Note that Carlini *et al.* [10] also suggest running the optimization algorithm multiple times with different random starting points (i.e., line 1) to avoid local optima. We employ the same trick and pick the best adversarial example generated among different executions of Algorithm 1 as the final result.

### A.4. Adversarial example generation algorithms details

In our evaluation, we examine both attack methods, i.e. Carlini *et al.* [10] and our proposed algorithm. For Carlini's attack, we choose to minimize the loss function:

$$\text{ReLU}(\mathcal{L}(J_\theta(x, Q^{\text{target}}), A^{\text{predict}}) \\ - \mathcal{L}(J_\theta(x, Q^{\text{target}}), A^{\text{target}})) + \lambda d(x, I^{\text{ori}})$$

where $x = 255 \times (\textbf{tanh}(\delta) + 1)/2$ to simulate the boxed constraint that each pixel value can only take value from $[0, 255]$. This approach is demonstrated to be the most effective one in [10]. Here $\lambda$ is chosen to be 0.1 by a grid search.

For our approach, as we discussed in Section 3, we choose the values of hyper-parameters as follows: $\epsilon = 2, \lambda_1 = 1, \lambda_2 = 10, \eta = 1.0, \textbf{maxitr} = 1000$. Note that these hyper-parameters are set based on each image being represented as a vector of pixel values from $[0, 255]$.

When we generate adversarial examples, we employ the RMSE distance function as used in [43]. In particular, assuming there are two $N$-dimensional vectors $x_1, x_2$, then the RMSE between the two vectors is computed as

$$RMSE(x_1, x_2) = \sqrt{||x_1 - x_2||_2^2 / N} = ||x_1 - x_2||_2 / \sqrt{N}$$

where $||\cdot||_2$ denotes the L2-norm of a vector. Further, in all experiments, the bound on the distance $B = 20$. In our experiments, the average distance for generated adversarial examples is below 10. We demonstrate several adversarial examples in Section **??** to illustrate that the generated adversarial examples are visually similar to their benign counterparts.

### A.5. Proofs to the theorems

We now present the formal proofs to Theorem 1 and Theorem A.2 presented in A.2.

*Proof of Theorem 1.* We consider two cases between the relationship between $A^{\mathbf{target}}$ and $A^{\mathbf{predict}}$. First, when $A^{\mathbf{target}} = A^{\mathbf{predict}}$, the left-hand side of (8) is 0, and thus (8) is trivially true.

Second, when $A^{\mathbf{target}} \neq A^{\mathbf{predict}}$, then the left-hand side of (8) becomes

$$\tau - \mathcal{L}(J_\theta(x, Q^{\mathbf{target}}), A^{\mathbf{predict}})$$

Thus proving (8) is equivalent to prove

$$\mathcal{L}(J_\theta(x, Q^{\mathbf{target}}), A^{\mathbf{predict}}) \leq \tau = \log K$$

We prove this by contradiction. Assuming $\mathcal{L}(J_\theta(x, Q^{\mathbf{target}}), A^{\mathbf{predict}}) > \log K$, then we have

$$-\log J_\theta(x, Q^{\mathbf{target}})_{A^{\mathbf{predict}}} > \log K$$

and thus

$$J_\theta(x, Q^{\mathbf{target}})_{A^{\mathbf{predict}}} < 1/K$$

By definition, we have

$$A^{\mathbf{predict}} = \mathbf{argmax}_i J_\theta(x, Q^{\mathbf{target}})_i$$

thus we know

$$\forall i \in \{1, ..., K\}. J_\theta(x, Q^{\mathbf{target}})_i < 1/K$$

Therefore, we know that

$$\sum_{i=1}^{K} J_\theta(x, Q^{\mathbf{target}})_i < K \times (1/K) = 1$$

However, we assume the last layer of $J$ is a softmax layer, and thus we have

$$\sum_{i=1}^{K} J_\theta(x, Q^{\mathbf{target}})_i = 1$$

which is a contradiction. Therefore, we conclude that Theorem 1 is true. □

*Proof of Theorem A.2.* We prove this by contradiction. We assume an adversarial example $I^{\mathbf{adv}} = I^\star$ does not satisfy (6), but optimizes (7). In this case, $d(I^{\mathbf{adv}}, I^{\mathbf{ori}}) > B > B - \epsilon$, and thus the ReLU function is activated and its output must be greater than $\epsilon$. Thus, the third component is at least $\lambda_2 \epsilon$. Since the other two components are also non-negative, therefore, the objective of (7) is at least $\lambda_2 \epsilon$ as well. On the other hand, we can set $I^{\mathbf{adv}} = I^{\mathbf{ori}}$, so that the value of objective (7) is at most $\mathcal{L}(f_\theta(I^{\mathbf{ori}}, Q^{\mathbf{target}}), A^{\mathbf{target}}) + \lambda_1 \tau$. Since $\lambda_2 \epsilon > \mathcal{L}(f_\theta(I^{\mathbf{ori}}, Q^{\mathbf{target}}), A^{\mathbf{target}}) + \lambda_1 \tau$, we have that setting $I^{\mathbf{adv}} = I^{\mathbf{ori}}$ results in a lower value of objective (7) than $I^{\mathbf{adv}} = I^\star$, which contradicts the assumption! □

## B. Further evaluation on DenseCap

We present the top-$K$ accuracy results for Caption B in Figure 7. The 17 failed adversarial examples of Caption A are presented in Figure 9. We omit the 148 failed adversarial examples of Caption B due to size limitation.

We also report the top-K accuracy results for our Gold set in Figure 8.

## C. VQA attack dataset construction details

We construct multiple attack datasets in our experiments. Each dataset contains a set of $(I, Q, A)$ triples, where $I$ is a benign image, and $(Q, A)$ is a target question-answer pair. We explain how these triples are selected in different datasets below.

**Gold.** For this dataset, we manually create triples where the target question is meaningful to the image, and the target answer is incorrect to the question and image pairs. To achieve this goal, we randomly select 100 images. For each of them, we manually choose questions that are meaningful to the image, while both MCB and N2NMN models can answer correctly the questions based on the image. If none of such questions exist for an image, we replace it with another randomly selected image. We repeat this process until we get 100 question-image pairs where both models predict correct answers. Then, for each question-image pair, we manually choose an answer that makes sense for the question but is incorrect in the context of the image. In the end, we have 100 $(I, Q, A)$ triples that constitute the Gold set.

**VQA-A.** This dataset is designed to be a combination of two sub-datasets, i.e., **Popular-QA** and **Rare-QA**. These two aim at evaluating the resilience of the two VQA models against adversarial examples with different target question-answer pairs.

For the **Popular-QA** dataset, we select 3,000 *popular* question-answer pairs. In particular, we first remove all answers appearing less than 3 times along with their questions in the VQA training set. This is because we observe that among these least frequent answers, many are simply typos,

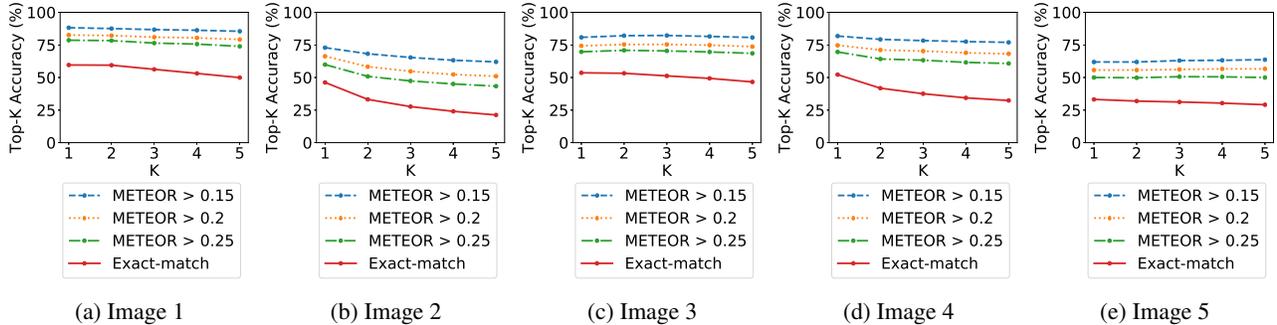

(a) Image 1    (b) Image 2    (c) Image 3    (d) Image 4    (e) Image 5

Figure 7: Top-$K$ accuracy on the Caption B dataset averaged across 5 images generated with each target caption

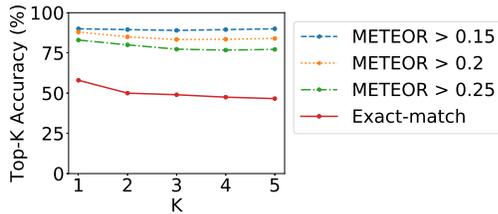

Figure 8: The result for adversarial attack against DenseCap model on the Gold set.

| Popular question-answer pairs | | |
|---|---|---|
| QA1 | What room is this? | bathroom |
| QA2 | What sport is this? | baseball |
| QA3 | What is the man doing? | skateboarding |
| QA4 | What is the man holding? | frisbee |
| QA5 | Is it raining? | no |
| Rare question-answer pairs | | |
| QA1 | What vegetable can be seen? | carrot |
| QA2 | What is the fence covered with? | net |
| QA3 | What does the blue signs represent? | handicap |
| QA4 | Why is the girl standing in the middle of the room with an object in each hand? | playing wii |
| QA5 | Who manufactured this plane? | japan |

Table 3: The question-answer pairs used in **Scale-Image**, popular (top) and rare (bottom).

(e.g., spelling "kitchen" as "kitten"). Therefore, we remove them from consideration.

We consider the top-1000 most frequent questions in the remaining set as *popular*. Further, for each popular question, we choose its top-3 most frequent answers and consider each corresponding question-answer pair as *popular*. To ensure each question has at least 3 answers, we also remove all questions with less than 3 answers before selecting the top-1000 most frequent questions.

We are interested in the popular question-answer pairs, because they appear more frequently in the training set, and thus the models may more likely remember these question-answer pairs. Therefore, we hypothesize that it is more likely to successfully generate an adversarial example with such a target for an irrelevant image. We create this dataset to test this hypothesis.

We also randomly select 5 images, which are provided in the top raw of Figure 11. For each question-answer pair $(Q, A)$ and each image $I$, we add the triple $(I, Q, A)$ to the Popular-QA set. In the end, there are 15,000 triples in this dataset.

The second dataset, i.e., **Rare-QA**, is similar to Popular-QA, but the question-answer pairs are *rare*. In particular, we filter out the answers appearing less than 3 times, and all questions with less than 3 remaining answers in the same way as during construction of Popular-QA.

Among the remaining questions, we select the top-1000 least frequent ones, and for each of them, we select the three least frequent answers. We consider the question-answer pairs selected by such criteria as *rare*, and in the end, we have 3,000 rare question-answer pairs. We use the same 5 images as in Popular-QA, and generate a triple using each question-answer pair and each image to construct 15,000 triples which constitute Rare-QA.

In doing so, we can evaluate the resilience of the two VQA models against adversarial examples on both popular question-answer pairs and rare question-answer pairs.

**VQA-B.** This dataset is similar to VQA-A, but is designed to evaluate the adversarial generation algorithm's performance across different benign images. To this end, we randomly select five popular question-answer pairs and five rare question-answer pairs, listed in Table 3, as well as 5,000 images to construct 50,000 triples in total. These triples constitute Scale-Image.

## D. More Results on Experiments with VQA

We present the CDF curves of adversarial examples generated using both our approach and Carlini's approach against MCB and N2NMN from the five images used in VQA-A, but we separately plot the analysis for Popular-QA and Rare-QA. For each combination of attack-model-dataset, we plot five curves on the same figure. The results are presented in Figure 10. The results show that for the same specification, the CDF curves for different images are close to each other. This shows that the attack performance is less dependent on images and more on the QA targets. We also see that the Rare-QA targets are more difficult than Popular-QA targets, and that our attack achieves higher probabilities that the Carlini's attack.

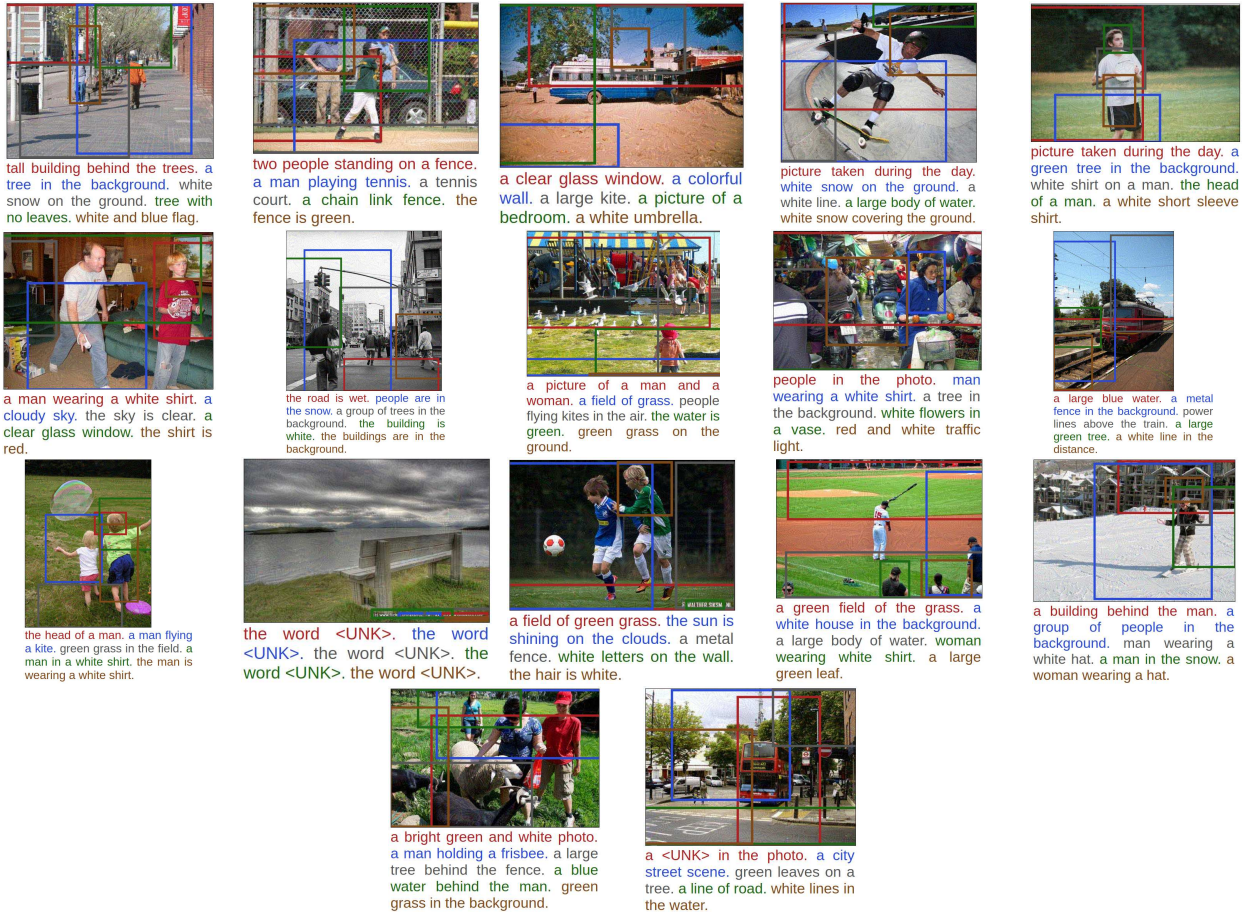

Figure 9: 17 failing adversarial examples generated from Caption A. For all these examples, the target caption is "white clouds in blue sky".

### D.1. Qualitative study

Figure 11 presents some qualitative examples from our experiments on the Rare-QA pairs. We provide both benign images and adversarial examples generated against MCB and N2NMN. We observe that it is hard to distinguish the benign images from adversarial ones visually.

We show the highest predictions of both VQA models on the benign images (top) and on the adversarial examples generated for the target QA pairs (bottom). We show targets in "[]" and highlight the failed attacks in *italics*. First, we note that even for the questions irrelevant to the images, initially, both VQA models can make reasonable predictions. We then review the models' behavior on the adversarial examples. We observe that the MCB model is more frequently fooled by the adversaries than the N2NMN model, for instance in the case of the first question. For the second question both models predict "left" instead of the target "to left", so essentially the attack succeeds, but it is counted as a failure case. Therefore, our quantitative results provide an over-conservative estimation on the attack success rate. Finally, for the third question all the attacks fail, and top predictions such as "yes" indicate the models' confusion. Interestingly, N2NMN model predicts "military" instead of "navy" for Image 2, which can also be counted as a success.

### D.2. Transferability Discussion

In this work, we focus on *white-box* adversarial examples, which means that the generation of these adversar-

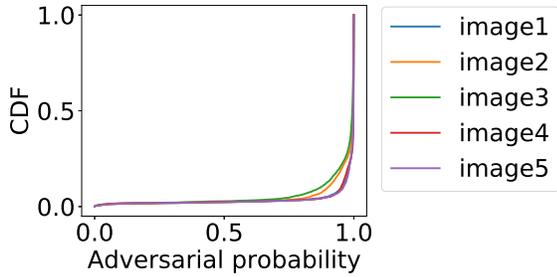
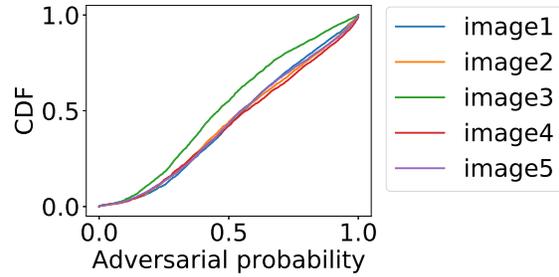

(a) CDF on adversarial probability of adversarial examples generated by our approach against MCB on Popular-QA.

(b) CDF on adversarial probability of adversarial examples generated by Carlini's approach against MCB on Popular-QA.

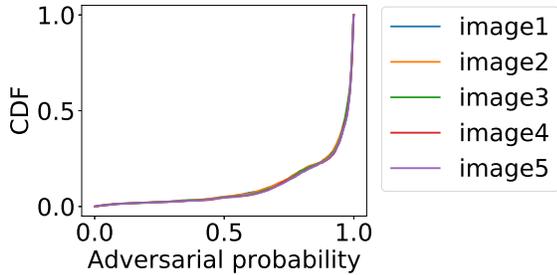
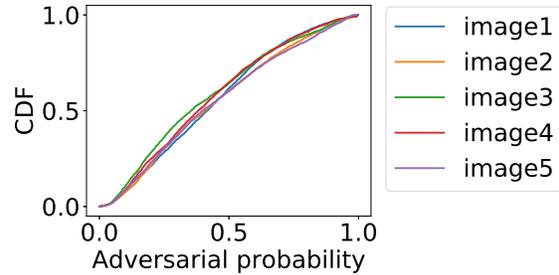

(c) CDF on adversarial probability of adversarial examples generated by our approach against N2NMN on Popular-QA.

(d) CDF on adversarial probability of adversarial examples generated by Carlini's approach against N2NMN on Popular-QA.

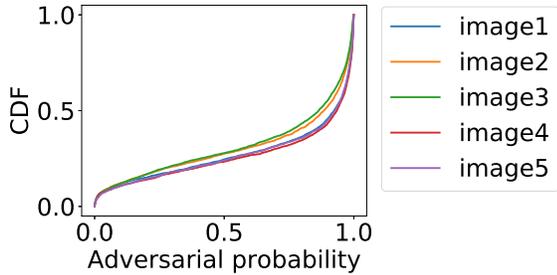
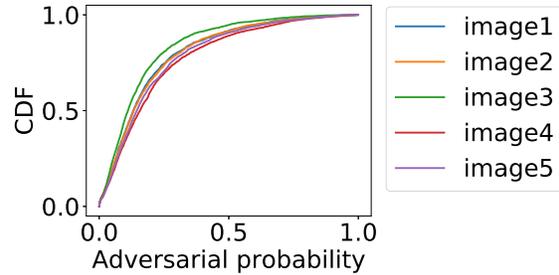

(e) CDF on adversarial probability of adversarial examples generated by our approach against MCB on Rare-QA.

(f) CDF on adversarial probability of adversarial examples generated by Carlini's approach against MCB on Rare-QA.

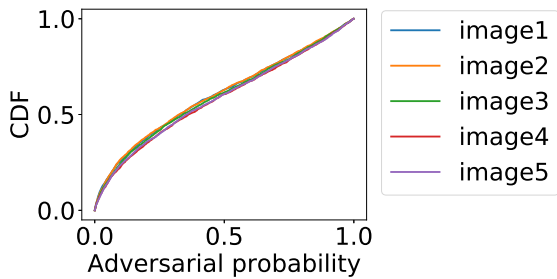
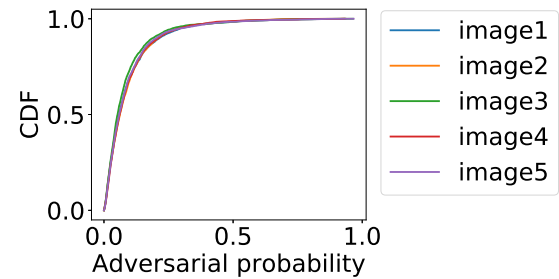

(g) CDF on adversarial probability of adversarial examples generated by our approach against N2NMN on Rare-QA.

(h) CDF on adversarial probability of adversarial examples generated by Carlini's approach against N2NMN on Rare-QA.

Figure 10: More CDF figures

ial examples requires full knowledge of the model architectures. However, we also demonstrate that an adversary could likely generate *black-box* adversarial examples without such knowledge. This is possible due to the *transferability* of adversarial examples, i.e., their ability to *transfer* between different network architectures [43, 53, 56, 66].

Previous work demonstrates transferability between: (1) two models with the same architecture trained on different

**Predictions on the benign images**

| | Image 1 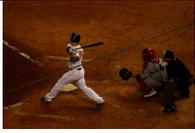 | Image 2 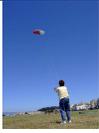 | Image 3 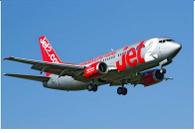 | Image 4 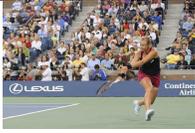 | Image 5 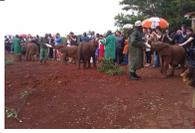 |
|---|---|---|---|---|---|
| | | | What are the people there for? | | |
| MCB: | baseball | flying kites | airplane | tennis | elephants |
| N2NMN: | baseball | kites | flying | tennis | parade |
| | | | Where is the boy's shadow? | | |
| MCB: | ground | ground | plane | court | ground |
| N2NMN: | ground | kite | sky | tennis court | ground |
| | | | Why is the man wearing a head covering? | | |
| MCB: | protection | safety | safety | tennis | protection |
| N2NMN: | protection | flying kite | safety | sweat | shade |

**Predictions on the adversarial examples**

What are the people there for? [festival]

| MCB: | festival | festival | festival | festival | festival |
|---|---|---|---|---|---|
| | 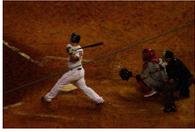 | 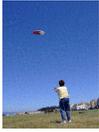 | 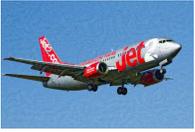 | 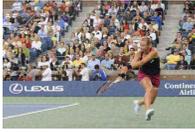 | 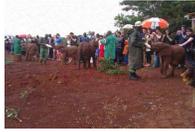 |
| N2NMN: | *parade* | *parade* | *parade* | *parade* | festival |
| | 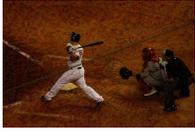 | 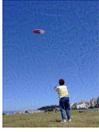 | 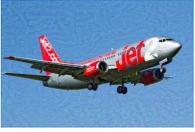 | 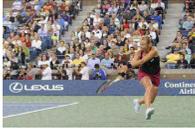 | 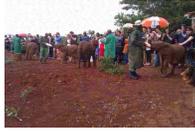 |

Where is the boy's shadow? [to left]

| MCB: | *left* | *left* | *left* | *left* | *left* |
|---|---|---|---|---|---|
| | 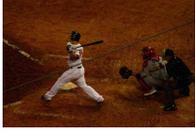 | 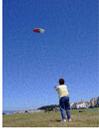 | 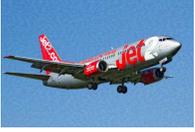 | 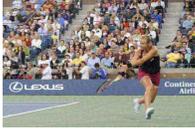 | 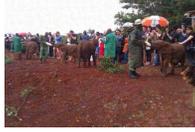 |
| N2NMN: | *left* | *left* | *left* | *left* | *left* |
| | 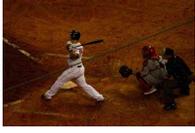 | 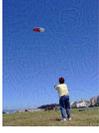 | 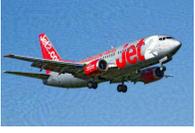 | 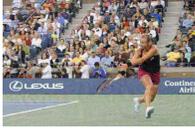 | 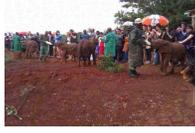 |

Why is the man wearing a head covering? [navy]

| MCB: | *yes* | *yes* | *safety* | *yes* | *costume* |
|---|---|---|---|---|---|
| | 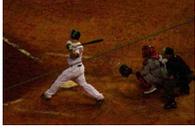 | 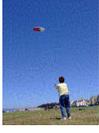 | 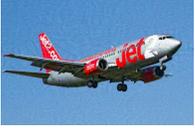 | 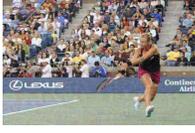 | 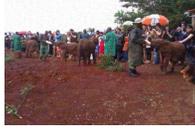 |
| N2NMN: | *yes* | *military* | *yes* | *yes* | *yes* |
| | 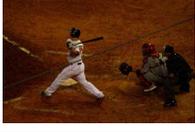 | 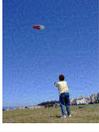 | 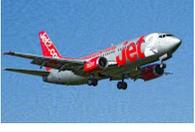 | 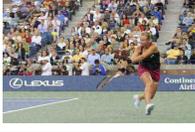 | 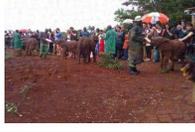 |

Figure 11: MCB and N2NMN's predictions on benign and adversarial images and QA pairs from **Rare-QA**. The target answer is provided in "[]" along with the question. The *text in italics* indicates that the targeted adversarial examples do not mislead the model to produce the exact target answer.

training data; (2) two models with different architectures trained on the same training data; and (3) even a neural network model and a non-neural network model (e.g., kNN, SVM). Most previous work demonstrates transferability of non-targeted adversarial examples. In [43], Liu *et al.* further demonstrate that almost none of targeted adversarial examples generated for one model transfer to another one, and developed a novel approach to generate targeted adversarial examples for an ensemble of multiple state-of-the-art classification models to achieve better transferability.

The transferability of adversarial examples enables the adversary to generate black-box adversarial examples from a white-box adversary. To do so, the adversary can simply generate adversarial examples against a white-box model that performs the same task as the black-box model, and these adversarial examples would transfer to the black-box model with a high probability. Papernot *et al.* show that they can effectively generate non-targeted black-box adversarial examples against black-box online machine learning systems hosted by Amazon, Google and MetaMind [57, 56]. Further, Liu *et al.* demonstrate successful non-targeted and targeted black-box adversarial examples against Clarifai.com, which is a commercial company providing state-of-the-art image classification services [43].

Again, all these previous work only study image classification models. In this work, we are interested in the transferability of targeted adversarial examples between vision-language models, which we show below.

**Experiment Results.** We test the transferability of the generated adversarial examples between MCB and N2NMN. We use the **Gold** set to generate adversarial examples for this evaluation. We find that 79 out of 100 adversarial examples generated for the MCB model can transfer to N2NMN, while the number is 60 in the other direction. This shows that adversarial examples on VQA models can transfer well, and thus opens the door for black-box attacks.

Notice that in existing work [43], Liu *et al.* demonstrate that it is non-trivial to generate transferable targeted adversarial examples from a single image classification model. We note that both MCB and N2NMN employ the same pre-trained ResNet-152 features [22] as their image representation. Thus, we attribute the good transferability results to the use of ResNet-152 in both models.

# E. Analysis on Hard Targets for Generating Adversarial Examples

While we observe that adversarial examples can be generated for most target question-answer pairs, in some cases the adversarial generation algorithm fails. We notice that whether the attack will succeed or not depends on the target question-answer pair rather than on the benign image. In this section, we investigate the failure cases and provide some insights into why some targeted attacks may be hard.

## E.1. The effectiveness of language priors

As we have observed in the experimental results described in Section 5 (in the main paper), whether a question-answer pair is a hard target depends more on the question-answer pair itself and less on the image. Therefore, we hypothesize that the language component in the VQA models may prevent adversarial examples to fool the models with certain targets. This phenomenon can be considered the *language prior* of VQA models. That is, given a question, if the model is less likely to predict a certain answer, we are also less likely to successfully generate targeted adversarial examples using it as the target answer.

In this section, we evaluate this phenomenon to verify our hypothesis. In particular, we choose a question, "What sport is this?". We first evaluate the **answer frequency** as follows. We run the VQA model on each of the 5,000 images in the VQA validation set and the selected question to get 5,000 answers. We compute the frequency of each answer in this set.

Intuitively, the answer frequency is a Monte-Carlo simulation of the answer distribution of the VQA model, and our goal is to examine the relationship between the answer distribution and the success of using an answer as the target to generate adversarial examples. In particular, we want to show that the answer frequency is positively correlated with the adversarial probability for each answer. To this end, we sequentially set each answer as the target answer, while setting the question chosen above (i.e., "what sport is this") as the target question, and Image 1 in Figure 11 as the benign image. Then we compute the adversarial probability of each answer. We sort all the answers in the descending order of their adversarial probabilities, and jointly plot the adversarial probabilities and the answer frequencies. Figures 12a and 12b show the corresponding plots for MCB and N2NMN. In these plots, each point in the x-axis indicates a label of an answer, so that the answer with the highest adversarial probability is labeled as 0, and so on. The blue line plots the adversarial probability of all answers, while the red dots plot the answer frequency. We only plot the answers whose frequency is at least 1, namely the answers must appear in the model's prediction set.

From both figures, we can observe a clear relationship between the answer frequency and the adversarial probability. That is, all answers with a frequency of 1 and higher can be predicted with a large probability (e.g., $> 0.1$), and all these answers can be used as targets to generate adversarial examples. Further, we observe that the answer frequency loosely aligns with the adversarial probability. This observation supports our hypothesis that the answer frequency is positively correlated with the adversarial probability.

Further, we observe that N2NMN has fewer answers with a positive frequency. We illustrate this phenomenon in Figure 12c. In this figure, we sort all answers in the descending order of their frequencies based on the N2NMN model, and the x-axis corresponds to their rank. The blue plot shows the distribution of the answer frequency computed based on the N2NMN model, while the orange dots are each answer's frequency computed based on the MCB model. We can observe that many answers have a large frequency based on the MCB model, but their frequency based on the N2NMN model is 0. Therefore, combined with the observation above, this demonstrates that the N2NMN model has a smaller range of answers that can be used as the target to generate adversarial examples than the MCB model.

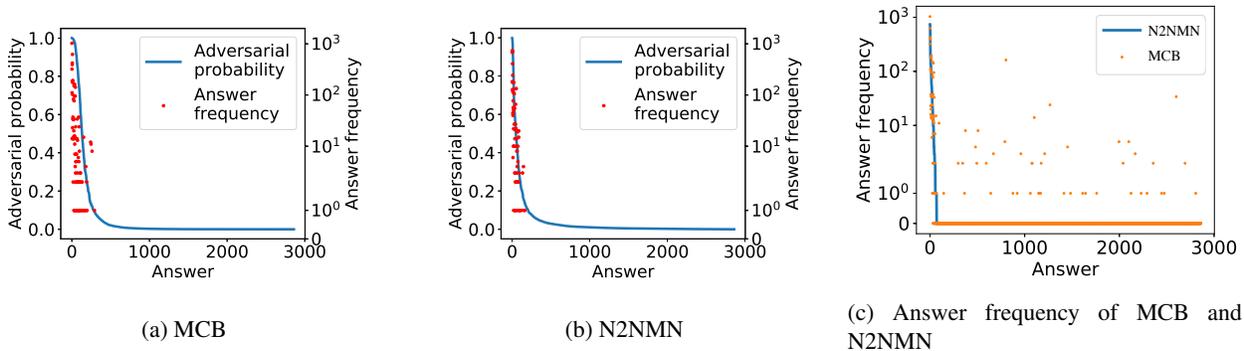

(a) MCB        (b) N2NMN        (c) Answer frequency of MCB and N2NMN

Figure 12: Answer frequency versus adversarial probability. Figure 12a and Figure 12b show that the answer frequency is positively correlated with the adversarial probability on MCB and N2NMN respectively. Figure 12c shows the answer frequency of the MCB model and the N2NMN model.

Notice that all these answers are generated based on the same questions. We investigate the results, and find that many of the answers predicted by the MCB model are irrelevant to the question used in this evaluation. This shows that, since N2NMN composes the network modules according to the input question, it is more effective at constructing corresponding filter modules, which can eliminate the answers irrelevant to the question. On the other hand, the MCB model does not have this functionality, since its architecture is identical throughout all questions. Therefore, when an image is less relevant to the question, the MCB model may predict answers considering the image more than the question. In this sense, the answer set of N2NMN is smaller than the one of MCB, since the former only includes answers relevant to the question. This also indicates that N2NMN has a stronger language prior than MCB, which partially explains why N2NMN behaves slightly more resilient than MCB in our previous experiments.

### E.2. Meaningless question-answer targets

We further evaluate the effect of language prior by constructing a dataset of *meaningless* question-answer targets.

We select 100 questions from 5 categories starting with (1) "What color"; (2) "What animal"; (3) "Is"; (4) "How many"; and (5) "Where". Then we construct the set of meaningful answers to each type of questions: for example, "silver" is a meaningful answer to a "what color" question. In doing so, the answer assigned to one type of question is guaranteed to be meaningless to the questions in another type. Thus we choose a meaningless answer for each of the 100 questions. We use them as targets and the 5 images used in Popular-QA and Rare-QA as the benign images to generate the adversarial examples. In the end, we observe that the attack success rates using our approach against MCB and N2NMN are only $7.8\%$ and $4.6\%$ respectively; the corresponding numbers for Carlini's attack are $6.8\%$ and $3.8\%$ respectively. This experiment further confirms the significance of the language prior and again demonstrates that N2NMN is more resilient against adversarial examples than MCB.